\documentclass[letterpaper, 10 pt, conference]{ieeeconf}  

\IEEEoverridecommandlockouts                              

\usepackage{times}
\usepackage{epsfig}
\usepackage{graphicx}
\graphicspath{ {figures/} }
\usepackage{subcaption}

\usepackage{amsmath}
\usepackage{amssymb}
\usepackage{multirow}
\usepackage{xspace}
\usepackage{txfonts}
\usepackage{textcomp} 

\usepackage{algorithm}
\usepackage{algcompatible}

\usepackage{url}

\algblockdefx{FORALLCUDA}{ENDFACUDA}[1]%
  {\textbf{for all }#1 \textbf{do on GPU cores in parallel}}%
  {\textbf{end for}}
  
\algblockdefx{FORCUDA}{ENDFORCUDA}[1]%
  {\textbf{for }#1 \textbf{do on GPU cores in parallel}}%
  {\textbf{end for}}

\makeatletter
\DeclareRobustCommand\onedot{\futurelet\@let@token\@onedot}
\def\@onedot{\ifx\@let@token.\else.\null\fi\xspace}

\def\eg{\emph{e.g}\onedot} 
\def\ie{\emph{i.e}\onedot}

\def\etal{\emph{et al}\onedot}

\newcommand{\bR}{\mathbf{R}}
\newcommand{\bt}{\mathbf{t}}

\newcommand{\bx}{\mathbf{x}}

\newcommand{\bP}{\mathbf{P}}
\newcommand{\bQ}{\mathbf{Q}}
\newcommand{\bp}{\mathbf{p}}
\newcommand{\bq}{\mathbf{q}}
\newcommand{\bT}{\mathbf{T}}
\newcommand{\bX}{\mathbf{X}}
\newcommand{\bn}{\mathbf{n}}

\newcommand{\bF}{\mathbf{F}}
\newcommand{\bI}{\mathbf{I}}
\newcommand{\bG}{\mathbf{G}}
\newcommand{\bK}{\mathbf{K}}

\title{\LARGE \bf
LoopSmart: Smart Visual SLAM Through Surface Loop Closure 
}

\author{Guoxiang Zhang$^{1}$ and YangQuan Chen$^{2}$
\thanks{This work was supported by CITRIS seed grant titled SmartCaveDrone, NVIDIA GPU grant and UC Merced Bobcat award.}
\thanks{$^{1}$Mechatronics, Embedded Systems and Automation Lab, School of
Engineering, University of California, Merced, Merced, CA, USA,
        {\tt\small gzhang8@ucmerced.edu}}%
\thanks{$^{2}$Mechatronics, Embedded Systems and Automation Lab, School of
Engineering, University of California, Merced, Merced, CA, USA,
        {\tt\small ychen53@ucmerced.edu}}%
}

\begin{document}

\maketitle
\thispagestyle{empty}
\pagestyle{empty}

\begin{abstract}

We present a visual simultaneous localization and mapping (SLAM) framework of
closing surface loops. It combines both sparse feature matching and dense
surface alignment. Sparse feature matching is used for visual odometry and
globally camera pose fine-tuning when dense loops are detected, while dense
surface alignment is the way of closing large loops and solving surface
mismatching problem. To achieve smart dense surface loop closure, a highly
efficient CUDA-based global point cloud registration method and a map content
dependent loop verification method are proposed. We run extensive experiments on
different datasets, our method outperforms state-of-the-art ones in terms of both camera trajectory and surface reconstruction accuracy.


\end{abstract}

\section{INTRODUCTION}

Our goal is to track a RGB-Depth (RGB-D) camera accurately in real time and optimize its trajectory when new information is available. Along with recovering camera trajectory, we also want to get globally consistent 3D models. This falls into the research area of visual simultaneous localization and mapping (SLAM). With both accurately estimated camera pose and 3D dense map, an SLAM system is very useful for a broad spectrum of applications, such as robot navigation, 3D dataset generation~\cite{xiao2013sun3d, dai2017scannet}, digital heritage~\cite{gzhangicuas2017} and augmented reality. 

There are still many challenges in making a well working SLAM system which emphasizes both camera trajectory and dense surface estimate accuracy, even though it has been actively researched for many years~\cite{Cadena16tro-SLAMfuture}. Major difficulties are from three sources. First, the camera tracking error accumulation problem can be considered unavoidable due to the incremental nature of any SLAM system. Second, it is hard to find all important loops. Intuitively, the more loops in data, the more information can be used to recover precise camera trajectory and the 3D model. But, in practice, when running existing SLAM systems on datasets with loopy motion at different scales, mismatches can always be found when scanned more than once. This means that loops are not successfully detected, indicating that the recovered camera trajectory is not precise enough. Third, an effective and optimal way of optimizing loop closure in a dense visual SLAM system is not yet attempted, due to perhaps the cost of reintegrating dense models.

To get a better camera tracking and dense mapping accuracy, researchers tried different ways to attack this problem. For a better tracking accuracy, in~\cite{7807268}, curvature information was added into the frame-to-model iterative closest point (ICP). To better close those loops, most SLAM systems~\cite{Whelan12rssw,endres20143,mur_artal_orb_slam2} use bag of words (BoW)~\cite{GalvezTRO12}, but it is well known that it is not very reliable under certain lighting condition or viewing angle changes. 
Since the BoW only matches sparse features from images but cannot fully utilize all camera observation data and spatial priors. On the other hand, SLAM systems tend to add the loops very conservatively so as to reduce severe influence of the false loops, thus many important loops may not be connected. Even after loops are successfully detected, there is still another problem in the dense SLAM system: how to correct reconstructed surface optimally. Since most dense visual SLAM systems~\cite{newcombe2011kinectfusion,Whelan12rssw,whelan2015elasticfusion} use a frame-to-model fusion process, which makes it difficult to quantify, isolate and remove the influence of past camera data, and it is also computational expensive for a full camera data sequence re-fusion. \cite{whelan2015elasticfusion} suggested to form a deformation graph across the reconstructed dense model to deform its surface to connect the loop. When loops become large, the model may not be deformed optimally, since past camera observation are not reused to manipulate the 3D model. They assumed that the scenes are elastic, but in reality, they are mostly rigid.


Motivated by the fact that human can notice mismatches in dense 3D models very easily by just looking at the spatial displacement of surfaces. We propose to resolve mismatches directly by closing dense loops to get a consistent 3D model and a precise camera trajectory estimate in the visual SLAM system. 
After dense loops detected, instead of optimizing dense surface directly to propagate correction introduced by dense loop, in this paper, dense loops correction is done through sparse feature bundle adjustment, so that all the past camera poses can be corrected based on their observations. By running extensive experiments on different datasets, we observed that combining sparse feature with dense loop closure can produce state-of-the-art performance not achieved before. Not only 3D model gets improved, but also camera trajectory estimate becomes more accurate. This is because our framework can detect loops in dense surface domain and optimize loops in sparse feature domain. Note that our framework can detect dense loops, yet other other means of detecting loops can still be utilized. 


In the following, we summarize the key contributions of our method:
\begin{enumerate}
\item We are the first to correct dense loops with sparse feature based bundle adjustment in a visual SLAM system. We demonstrate this dense-sparse combination can give much improved camera tracking and dense modeling results.
\item We propose and implement a CUDA based global point cloud registration algorithm that is fast and accurate so that it can be used in a real-time SLAM system; 
\item We can correct dense surface in real time based on the motion of camera trajectory estimate. 
\end{enumerate}


\section{RELATED WORK}
\label{sec:rw}

Visual SLAM has been studied actively for many years. Researchers from different fields, such as robotics, computer vision and computer graphics, tried to solve this problem with their own emphases and preferences, which lead to diverse visual SLAM systems.

Sparse feature based SLAM systems are well developed, because sparse features can be used to sample data from sensor reading (\eg images) to sparse data representation as image keypoints and feature vectors, which means less computation, since data from different frames is matched solely based on feature vectors of their keypoints. Extended Kalman filter or particle filter based filtering approaches~\cite{990501,1241885} can take keypoints as visual landmarks and solve visual SLAM as a data filtering process. A drawback of this approach is that 
the filter cannot be re-optimized again based on all previous data. Then, maximum a posteriori (MAP) based approaches are used to optimize all observed camera data in a batch setting~\cite{mur_artal_orb_slam2}, which utilizes bundle adjustment (BA) from Structure from Motion~\cite{4209727, Dai_bundle_fusion}, to get a better accuracy. They usually have local BA in camera tracking to have map size independent time complexity~\cite{mur_artal_orb_slam2}. In order to run BA for loop closure, the loops need to be detected. In the sparse image feature setting, BoW based loop closure is widely used. But it can give high portion of false loops, which can severely degrade performance of a SLAM system, so a very strict loop filtering is often used~\cite{mur_artal_orb_slam2}, where many loops are blindly rejected. This causes a big problem when there are many loopy motion in camera movement. 


Another line of visual SLAM research tried to focus on dense surface reconstruction. With the parallel processing power of GPU, Newcombee \etal proposed  KinectFusion~\cite{newcombe2011kinectfusion} which performs real-time dense 3D camera tracking and model fusion. It has a volumetric scene representation, which can be rendered to a depth map at a given camera pose. Tacking is done through a frame-to-model projective ICP, which is parallelized on GPU for real-time performance. Finally, new camera data is fused into the volumetric model using running average. KinectFusion can be considered of fusing very local loops together using the model it maintains as a proxy, but it does not close large loops. To close large loops, how to detect loops and find relative poses between loop areas need to be solved. In BoW, image keypoints and feature are used for both loop detection and relative pose generation, but in dense 3D systems, there is no such comparably reliable point cloud feature. Whelan \etal use BoW in a dense SLAM system called Kintinuous~\cite{Whelan12rssw}, which is an extended KinectFusion system. Later, to better solve loop detection and optimization problem, ElasticFusion~\cite{whelan2015elasticfusion} proposed to use ICP to find relative poses of potential loops, which are proposed by two sources of information: spatial prior and appearance-based place recognition. Our work shares similarity with~\cite{whelan2015elasticfusion} in using the spatial prior, but we propose a totally different approach and underlying algorithms. Instead of using projective ICP which highly depend on initialization. We propose a GPU based global point cloud registration method to close loops with another prior information from sparse feature alignment. 
After dense loop detected, in Kintinous, a pose graph of keyframes was utilized, while the authors mentioned that mesh deformation was required to get smooth 3D models, which indicates loop correction is not done optimally. In ElasticFusion, the pose graph is totally replaced by a deformation graph distributed inside the dense model. This deformation graph does not have a backing physical meaning, because most of scenes scanned are not elastic. In our framework, we utilize bundle adjustment to have an MAP correction of all past keyframes, which is theoretically optimal. BundleFusion~\cite{Dai_bundle_fusion} used bundle adjustment to optimize loops, but they do not close dense loops, instead, they close sparse feature loops and only use dense surface for feature correspondence search and tracking. Also, they use the idea of re-integrating camera data, which consumes much more computation than our approach.

\section{FRAMEWORK OVERVIEW}
\label{sec:fw}
\begin{figure}[hb]
\centering
\includegraphics[width=8cm]{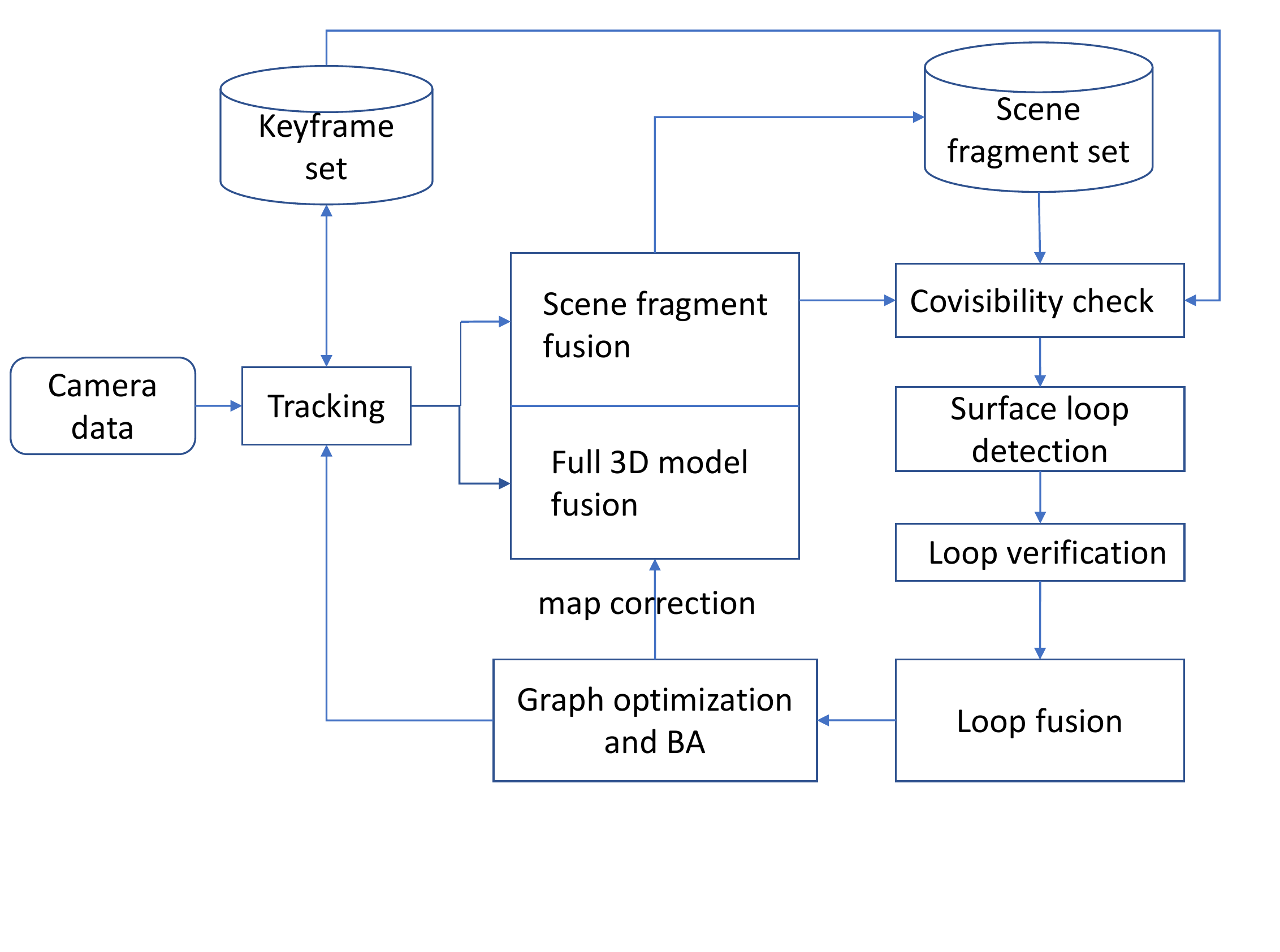}
\caption{System diagram}
\label{fig:system_diagram}
\end{figure}

We divide our framework to four major parts: 1) tracking, 2) dense model fusion, 3) dense loop detection and verification, 4) camera poses optimization and map correction. 
They are tightly coupled, as in Fig.~\ref{fig:system_diagram}, so that more information can be exploited out of sensor data. 

\textbf{Tracking.} We employ the tracking part from ORB-SLAM2 \cite{mur_artal_orb_slam2}, which is a very well implemented sparse feature based SLAM system.  Inside this tracking module, Oriented FAST and Rotated BRIEF (ORB) feature is extracted for keypoint matching. Then frames are tracked against keyframes with motion estimate and then refined with the local sparse map. Keyframes are generated when tracking is weak or local bundle adjustment thread is free. Local BA is used to correct re-projection error of feature correspondences among co-visible keyframes in a background thread.

\textbf{Dense model fusion.} We fuse dense models on a GPU using surfels as a map representation similar to \cite{keller2013real, whelan2015elasticfusion}. Each surfel has a position $\bp$, normal vector $\bn$, radius $r$, confidence $c$, initialization time stamp $t_0$, last updated time stamp $t_u$. It has a sparse data structure because it only stores data where surface exists, thus it is suitable for larger scene. More importantly, when the camera trajectory gets optimized, it is easier to manipulate a surfels model than a model with volumetric representation. 
We fuse every $k$ frames ($k=50$ for all experiments, as in \cite{Choi_2015_CVPR}) into a fragment of dense model, called scene fragment. 
These scene fragments are generated for two reasons. One is to integrate out raw RGB-D data noise. Another one is to reduce the number of 3D pieces so that loop detection computation are accelerated. After each scene fragment is generated, it is linked to the keyframe whose time stamp is closest to the $(\frac{k}{2})$-th frame, for the purpose that scene fragments can be transformed accordingly when camera trajectory estimate is updated.

\textbf{Dense loop detection and verification.} In order to close dense loops effectively and efficiently, we utilize strong prior and CUDA accelerated computing. First, only the scene fragment pair with overlapping in camera observation can be a loop candidate. Second, we use a co-visibility graph to pre-filter loop proposals that are co-visible. Then filtered loop proposals are aligned using a CUDA based global point cloud registration method we proposed. If two point cloud can be aligned together, the loop proposal becomes a loop candidate with a relative transformation matrix that can align them together. Finally, the loop candidates need to go through a loop verification process, so that false loops would not diverge the subsequent optimization process.

\textbf{Loop optimization and correction.} After dense loops detected and verified, they are used to connect pose graph and also trigger more image loop detection, which again uses spatial prior and ORB feature matching. Then, the pose graph is optimized to give a coarse pose correction and then a full BA is performed in order to get MAP optimally fine-turned camera trajectory estimate. Finally, both spare map and dense model are corrected based on the update of camera trajectory estimate.

\section{FAST GLOBAL POINT CLOUD REGISTRATION ON GPU}
\label{sec:fpcrg}

\begin{algorithm}[ht]
\caption{Global Registration on GPU}
\label{ag:reg}
 \hspace*{\algorithmicindent} \textbf{Input:} A pair of point cloud $\bP$ and $\bQ$ \\
 \hspace*{\algorithmicindent} \textbf{Output:} $\bT_{\bQ\bP}$ if $\bP$ and $\bQ$ can be aligned together;
\begin{algorithmic} 
\STATE Downsample $\bP$ and $\bQ$;
\STATE Compute FPFH features {$\bF(\bQ)$} and {$\bF(\bQ)$};
\STATE $\bT_{QP} \leftarrow \bI$;
\STATE  $feature\_NN\_cache \leftarrow \emptyset$ 

\STATE Quadruple point set samples $S_{P}  \leftarrow \emptyset$ , $S_{Q}  \leftarrow \emptyset$ 
\vspace{0.15cm}
\STATE //\textit{ Parallel FPFH feature pre-matching} 
\FORCUDA{$i \leftarrow 1$ \textbf{to} $\bP\textunderscore count$}
\STATE $feature\_NN\_cache[i] \leftarrow$  NN of $\bF(\bp_i)$ in {$\bF(\bQ)$};
\ENDFORCUDA

\vspace{0.15cm}
\STATE //\textit{ Randomly sample hypotheses on GPU} 
\FORCUDA{$i \leftarrow 1$ \textbf{to} $sample\textunderscore number$}
\STATE $S_{P}[i] \leftarrow$ randomly picked $(\bp_0 , \bp_1 , \bp_2 , \bp_3 )$ from $\bP$ ;
	\FORALL{$\:(\bp_{0} , \bp_1 , \bp_2 , \bp_3 )$}
		\STATE $\bq_j \leftarrow$ NN of $\bp_j$ in $\bQ$ using $feature\_NN\_cache[j]$
	\ENDFOR
\STATE $S_{Q}[i] \leftarrow (\bq_{0} , \bq_1 , \bq_2 , \bq_3 )$

\vspace{0.15cm}
\STATE //\textit{ Hypotheses pre-rejection}
  \FORALL{$\:(\bp_{0} , \bp_1 , \bp_2 , \bp_3 ), (\bq_{0} , \bq_1 , \bq_2 , \bq_3 )$}
  \IF{$||\bp_k - \bp_{k+1}|| < \tau ||\bq_k - \bq_{k+1}||$ or vice versa} 
      \STATE set current hypothesis pre-rejected in $S_{P}$ and $S_{Q}$;
  \ENDIF
  \ENDFOR
\ENDFORCUDA
\STATE Stream compact non-rejected hypotheses;

\vspace{0.15cm}
\STATE //\textit{ Consensus for the remaining quadruple pairs}
\FORCUDA{$i \leftarrow 1$ \textbf{to} $remain\_number$}
\STATE Estimate transformation $\bT_{\bQ\bP}$ from $S_{P}[i]$ to $S_{Q}[i]$ ;
\STATE Find all inliers under $\bT_{\bQ\bP}$ hypothesis;
\STATE  $test\_log \leftarrow$ {inlier ratio, fitness score}

\ENDFORCUDA

\vspace{0.15cm}
\STATE //\textit{ Get final result from test\_log}
\STATE Find the Hypothesis that has max inlier ratio and fitness score passing minimum threshold in $test\_log$.

\IF{Hypothesis found}
\STATE return its $\bT_{\bQ\bP}$;
\ENDIF

\end{algorithmic}
\end{algorithm}
To formulate dense loop closure formally, we denote a dense scene fragment as a set of points $\bP$ with their normal. Then the dense fragment based loop closure problem can be solved by point cloud registration methods. In a SLAM system, this registration needs to be done very quickly to meet the real-time criteria. In ElasticFusion, Whelan \etal resort to projective ICP, which can be performed very quickly on GPU. But, when mismatches are big, the initialization-dependent nature of ICP makes it difficult to converge to the right solution. So we turn to global point cloud registration in our framework, which does not depend on initial alignment at all.  From evaluation in \cite{zhou2016fast}, global registration methods are either too slow or have unreliable performance. We find the one \cite{Choi_2015_CVPR} with best recall can be parallelized and accelerated using GPU computing so that it can be both the most accurate and fastest.

\begin{table*}[bht]
\centering
\caption{Performance evaluation of different point cloud registration methods}
\label{tb:pcd_reg_eval}
\begin{tabular}{|l||c|c|c||c|c|c||c|c|c|}
\hline
\multirow{2}{*}{} & \multicolumn{3}{c||}{Time/pair (milliseconds)} & \multicolumn{3}{c||}{Recall}       & \multicolumn{3}{c|}{Precision}    \\ \cline{2-10} 
                  & CZK         & FGR        & Ours               & CZK    & FGR    & Ours            & CZK    & FGR             & Ours   \\ \hline
Living room 1     & 7606.20     & 131.94     & \textbf{24.71}     & 61.3\% & 53.1\% & \textbf{62.4\%} & 27.2\% & \textbf{34.9\%} & 23.6\% \\ \hline
Living room 2     & 7469.58     & 80.96      & \textbf{18.19}     & 49.7\% & 37.9\% & \textbf{50.3\%} & 17.0\% & \textbf{24.5\%} & 20.5\% \\ \hline
Office 1          & 7556.02     & 100.36     & \textbf{21.63}     & 64.4\% & 53.9\% & \textbf{67.8\%} & 19.2\% & \textbf{28.6\%} & 19.3\% \\ \hline
Office 2          & 7418.12     & 74.41      & \textbf{17.45}     & 61.5\% & 56.3\% & \textbf{71.1\%} & 14.8\% & \textbf{24.5\%} & 16.4\% \\ \hline
Average           & 7512.23     & 96.92      & \textbf{20.50}     & 59.2\% & 50.3\% & \textbf{62.9\%} & 19.6\% & \textbf{28.1\%} & 20.0\% \\ \hline
\end{tabular}
\end{table*}

Our algorithm is based on RANSAC and inspired by \cite{Choi_2015_CVPR}, which is modified from Point Cloud Library \cite{5152473}. The major differences are that we accelerated the most time consuming parts using GPU programing with an efficient nearest neighbor search method, and that normal checking is moved from pre-rejection part into hypotheses testing. Traditionally, RANSAC is formulated as an iterative process with proved convergence~\cite{Fischler:1981:RSC:358669.358692}. But, if we relax the theoretical convergence requirement, different iterations and different hypotheses can be considered to be totally independent to each other. This means different hypotheses can be mapped to different processing cores to be tested in parallel.

As shown in Algorithm \ref{ag:reg}, point clouds are downsampled to the resolution of the typical precision of RGB-D sensors to reduce unnecessary computation. Fast Point Feature Histograms (FPFH) features are extracted for each point in $\bP$ and $\bQ$ for point correspondence pairs generation. To make nearest neighbor search more efficient, we pre-cached all the nearest neighbors of $\bP$ in $\bQ$ using FPFH feature distance. Then, for each hypothesis, 4 points are randomly sampled from $\bP$, and their correspondences are found through the pre-cached nearest neighbors. After that, a pre-rejection step, which rejects hypotheses whose point pairs cannot make a similar polygon, is performed. $\tau$ is a similarity threshold and set to $0.9$ in all our experiments. Then, hypotheses testing, the most time consuming step, tests both inlier ratio and fitness score on GPU. When implement it, we test each hypothesis on a thread block with efficient parallel reductions. For the NN search during hypotheses testing, we utilized a 3D grid to replace the k-d tree to fit special need of a GPU, since a GPU will slow down when different threads go to different code branches during k-d tree search. We propose to use a 3D grid for NN search, given that the point clouds to be searched only span in a limited area. This guarantees that we can use a grid with limited size for NN searching without jeopardizing searching accuracy. When a point is stored into the search grid, the indices of its cell is calculated by eq.~(\ref{eq:index}).
\begin{equation}
\label{eq:index}
\begin{split}
index_x &= (x_p - x_{center})/cell\_length,\\
index_y &= (y_p - y_{center})/cell\_length,\\
index_z &= (z_p - z_{center})/cell\_length,
\end{split}
\end{equation}
where the $x_{center}, y_{center}, z_{center}$ are the coordinate of the center of target point cloud. $cell\_length$ is the cell edge length of the NN 3D grid. It is subtracted so that the translation of point cloud does not affect searching. 
When a point wants to query its nearest neighbor, the searching is accomplished through table looking up, which has a time complexity of $\mathcal{O}(1)$, given cell edge length the same as point cloud downsample resolution. It is faster than the k-d tree which has a $\mathcal{O}(\log{n})$ time complexity. We observed speedup by only replacing k-d tree with 3D search grid in CPU only code. More importantly, there is no branching during searching, so it fits much better on a GPU than the k-d tree.

We run experiments to compare speed, recall and precision performance against two baseline methods: CZK~\cite{Choi_2015_CVPR} and FastGlobalRegistration (FGR) \cite{zhou2016fast} on redwood pairwise registration evaluation dataset by Choi \etal. We report results in Table \ref{tb:pcd_reg_eval}. Recall and precision for CZK are from reports of paper authors, but speed is measured on our machine. Not all the complete results of FGR is publicly available from its authors, so we measure them on our own. For all methods, we use the same hyper-parameter values as in published code of~\cite{Choi_2015_CVPR}: $0.05$~m as point cloud downsampling leaf size, $0.1$~m as normal estimate radius, $0.25$~m as FPFH feature estimate radius, $4000000$ as hypotheses count and $0.075$~m as maximum point correspondence distance.  We use an Intel i7-6850K clocked at 3.6 GHz and an NVIDIA Titan X Pascal for our evaluation.

Our global point cloud registration can finish in around 20 milliseconds, which is around 366 times faster than CZK and 5 times faster than FGR as in Table \ref{tb:pcd_reg_eval}. With this speed, it can run at 50 Hz, which means we can process more loop candidates. Our recall performance is not compromised for speed, instead, it is even better than all these two methods. Even though it is much faster than other methods, it is necessary to figure out a smart way of using our point cloud registration method for a high performance SLAM system. So we propose ways to pre-filter unnecessary loop detection using spatial priors. From Table \ref{tb:pcd_reg_eval}, the low precision is an issue needs to be solved, so an effective loop verification method is also adapted. We describe these in the following section. 

\section{DENSE LOOP PROPOSAL AND POST-FILTERING}
\label{sec:looprej}

To avoid unnecessary loop detection computation, we use strong spatial priors and connections between dense model and sparse map to generate loop proposals and then only run global point cloud registration on these loop proposals. The first criteria of dense loop proposal is spatial overlapping of scene fragments pair in their camera observations. This is based on the prior that mismatch is more likely to happen when two surfaces have overlapping. One may argue that this will limit the scope of loops can be closed, but given indoor environment is usually small and tracking model from ORB-SLAM2 is considerably accurate, we do not observe any far away drift without any surface overlapping. If that could happen, we can easily solve it by adding image based place recognition to generate more spatial location independent loop proposals, which fits perfectly with the rest of the processing pipeline. 

There is more information can be exploited to filter these loop proposals, because it is not always necessary to close dense loops when sparse feature tracking itself has tracked new area into the old map. Or, in the case that, after closing one dense loop, sparse feature tracking successfully find sparse map connection between new map and old map, then there is no need to continue detecting dense loops. This makes our dense loop detection smart, so it only run detection when necessary.

All the loop proposals will be fed into GPU based point cloud registration to recover relative pose of aligning two point clouds, if they can be aligned. As mentioned before, the low precision of a global registration method is a serious problem, so we add a loop post-filtering process. Our idea is to utilize information of pose graph. Because when doing experiments, we find false loop usually bring in dramatic changes of pose graph structure, while true positive loops  make  just small  correction. Again, this assumption applies when the tracking model is reliable enough, which holds in our case.

To formulate this prior information mathematically, we turned to line processes based robust optimization proposed in \cite{Choi_2015_CVPR}. Our adapted objective function is eq. (\ref{eq:cvpr15}): 
\begin{equation}
  \label{eq:cvpr15}
  \begin{split}
    E(\mathbb{T}, l) &= \lambda\sum\limits_{s} f(\bT_s, \bT_{s+1}, \bT_{s,s+1})\\
    &+ l_{ij}f(\bT_i, \bT_j, \bT_{i,j}) + \mu\Psi(l_{ij}),
  \end{split}
\end{equation}
where
\begin{equation}
  \label{eq:cvpr15_f_org}
    f(\bT_i, \bT_j, \bT_{i,j}) = \sum\limits_{(\mathbf{p,q}) \in \mathcal{K}_{ij}} \left\Vert \mathbf{T}_i\mathbf{p} - \mathbf{T}_j\mathbf{q}\right\Vert ^2,
\end{equation}
\begin{equation}
  \label{eq:fast_l_multi}
  \Psi(l) = (\sqrt{l} - 1)^2,
\end{equation}
same as in \cite{Choi_2015_CVPR}, $\bT_i$ is the i-th camera pose in 4-by-4 matrix format, $\bP_i$ is the i-th point cloud, $\bT_{i,j}$ is a matrix that transform points from the coordinate system of $\bP_j$ to the coordinate system of $\bP_i$, $\mathcal{K}_{ij}$ is the set of nearest neighbor correspondence pairs between $\bT_i\bP_i$ and $\bT_j\bP_j$ that are within distance $\varepsilon$ = 0.05 m, which is typical noise level of RGB-D sensor. $\lambda$ is weighting factor of odometry term and is set to 1000 in all our experiments.  $\mu=\tau\kappa$, and $\tau=0.2$ is a distance threshold. Weight $l$ which controls the influence of the loop candidate tested, is optimized during iterations. To accelerate this computation, eq. (\ref{eq:cvpr15_approx}) is formulated to approximate eq.~(\ref{eq:cvpr15_f_org}) in \cite{Choi_2015_CVPR} :
\begin{equation}
  \label{eq:cvpr15_approx}
    f(\bT_i, \bT_j, \bT_{i,j}) = \xi^\intercal\left(\sum\limits_{(\mathbf{p,q}) \in \mathcal{K}_{ij}}\bG^\intercal\bG\right)\xi,
\end{equation}
where $\xi$ is a 6D rotation and translation vector $(\alpha, \beta, \gamma, x, y, z)$ to represent matrix $\bT_{i,j}\bT_{j}^{-1}\bT_{i}$, $\bG_p = \left[-[\bp]_{\times} | \bI\right]$. 
With this approximation, this objective function is solved using g2o. A loop proposal is rejected when $l < 0.25$. We periodically (every 50 fragments in all experiments) update $\bG^\intercal\bG$ so that our graph is not outdated.

\section{LOOP OPTIMIZATION AND CORRECTION}
\label{sec:map_opt}

After a dense loop passing verification, map optimization and correction are followed to reduce mismatches and error. We employ BA to get a MAP based optimization to correct the camera trajectory estimate and the dense model. During optimization, model area with earlier time stamp is fixed as a target. The rest model is moved to match it. First, the key frame associated with the source scene fragment get transformed into coordinate system of target scene fragment using the relative pose $\bT$. Its co-visible keyframes are also transfered using $\bT$. Then more image based loop detection is triggered to detect more loops. After detecting new image feature loops, similar to \cite{mur_artal_orb_slam2}, a pose graph optimization is performed to give a rough camera pose trajectory and then bundle adjustment is performed by optimizing eq.~(\ref{eq:ba}):
\begin{equation}
  \label{eq:ba}
   \text{min} \sum\limits_{c}\sum\limits_{p\in V(c)} \rho\left(\left\Vert\tilde{\bx}_p^c-\bK[\bR_c|\bt_c]\bX_p\right\Vert_{\Sigma} \right),
\end{equation}
where $\rho$ is the robust Huber cost function, $\tilde{\bx}_p^c$ is the observed 2D pixel location in the camera coordinate system, $\bK$ is the camera intrinsics matrix, $\bR_c$ and $\bt_c$ are the camera orientation and position of c-th keyframe pose $\bT_c$, $\bX_p$ is the location a point visible from the c-th camera,  $\Sigma$ is the covariance matrix associated to the scale of the keypoint.

Since we have a front end of displaying real-time dense surface. It is important to have dense model corrected in real time as well. Previously, \cite{Dai_bundle_fusion} uses an idea of de-fusion and reintegration on a volumetric representation, but, as mentioned before, de-fusion process cannot really reverse the fusion process, due to its filtering nature. In ElasticFusion, the surface is deformed based on a deformation graph. Here, we propose a novel way of correct dense surface based on the correction motion of camera pose estimate. The idea is to move the dense surface with observing keyframe cameras. When pose corrections are applied to a keyframe, we move all the surfels it can observe together with this correction. To apply this idea, two problems need to be dealt with: 1)~Most surfels are observed from adjacent camera poses; 2)~Surfels from overlapping area can be observed from two distinctly different camera sub-trajectory, \ie the trajectory in loop area. We solve the first problem by averaging the influence from adjacent keyframe camera poses. Our solution to the second one is to take the influence of keyframe set $\mathcal{F}$ only when $t_{0} < t_{keyframe} < t_{u}$, so that only keyframes in the time window between initial and updating are considered. The updated position and normal of a surfel is formulated as eq.~(\ref{eq:surfels_update})
\begin{equation}
  \label{eq:surfels_update}
  \begin{split}
   \bp_{updated} &= \frac{1}{k}\sum\limits_{f \in \mathcal{F}} \bT_{f\_new}\bT_{f\_old}^{-1} \bp_{old},\\
   \bn_{updated} &= normalize\left(\frac{1}{k}\sum\limits_{f \in \mathcal{F}} \bR_{f\_new}\bR_{f,old}^{-1} \bn_{old}\right).
  \end{split}
\end{equation}
We also remove points who are not observed from any camera pose. This dense model correction can be done very efficiently on GPU, since each surfel only take influence from very limited numbers of keyframes and there is no inter-surfel dependence.
\begin{table*}[ht]
\centering
\caption{Surface reconstruction error (in meters) on augmented ICL-NUIM sequences}
\label{tb:surface_error}
\begin{tabular}{|l||c|c|c|c|c|c|c|c|c|c|}
\hline
\multirow{2}{*}{} & \multicolumn{2}{c|}{Living room 1} & \multicolumn{2}{c|}{Living room 2} & \multicolumn{2}{c|}{Office 1}    & \multicolumn{2}{c|}{Office 2}    & \multicolumn{2}{c|}{Average}    \\ \cline{2-11} 
                  & mean            & median         & mean            & median         & mean           & median         & mean           & median         & mean           & median         \\ \hline 
ElasticFusion     & 0.084           & 0.050          & 0.111           & 0.062          & 0.081          & 0.036          & 0.043          & 0.019          & 0.080          & 0.042          \\ \hline
BundleFusion      & 0.061           & 0.020          & 0.065           & 0.022          & 0.016          & 0.053          & 0.030          & 0.019          & 0.043          & 0.029          \\ \hline
CZK               & 0.033           & 0.019          & 0.028           & 0.019          & 0.019          & 0.015          & 0.022          & 0.014          & 0.026          & 0.017          \\ \hline
ORB-SLAM odometry & 0.031           & 0.015          & 0.022           & 0.018          & 0.019          & 0.012          & 0.014          & 0.010          & 0.022          & 0.014          \\ \hline
ORB-SLAM full     & 0.017           & 0.010          & 0.010           & 0.008          & 0.015          & 0.010          & 0.013          & 0.010          & 0.014          & 0.010          \\ \hline
Ours              & \textbf{0.007}  & \textbf{0.005} & \textbf{0.007}  & \textbf{0.006} & \textbf{0.013} & \textbf{0.009} & \textbf{0.010} & \textbf{0.007} & \textbf{0.009} & \textbf{0.006} \\ \hline
\end{tabular}
\end{table*}
{\centering
\begin{figure*}[ht]
\begin{tabular}{cccc}
\begin{subfigure}{0.25\textwidth}\centering\includegraphics[width=1\linewidth]{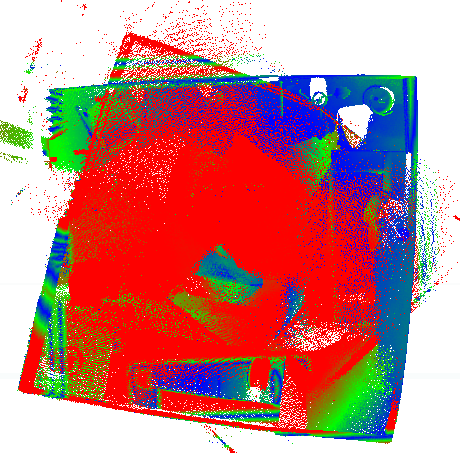}\caption{ElasticFusion}\label{fig:taba}\end{subfigure}&
\begin{subfigure}{0.25\textwidth}\centering\includegraphics[width=1\linewidth]{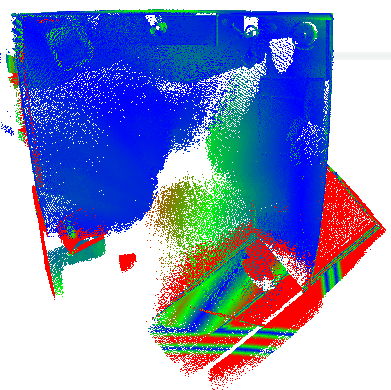}\caption{BundleFusion}\label{fig:taba}\end{subfigure}&
\begin{subfigure}{0.25\textwidth}\centering\includegraphics[width=1\linewidth]{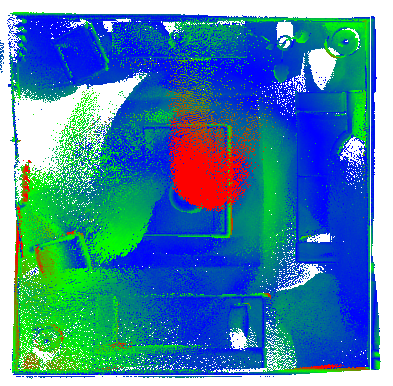}\caption{CZK}\label{fig:tabb}\end{subfigure} & \multirow{2}{*}{\begin{tabular}[c]{@{}l@{}}$\quad \: 0.01\:m$ \\ \begin{subfigure}{0.08\textwidth}\centering\includegraphics[width=0.33\linewidth]{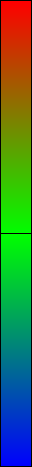}\label{fig:taba}\end{subfigure}\\ $\quad\quad 0 \: m$ \end{tabular}}\\
\newline
\begin{subfigure}{0.25\textwidth}\centering\includegraphics[width=1\linewidth]{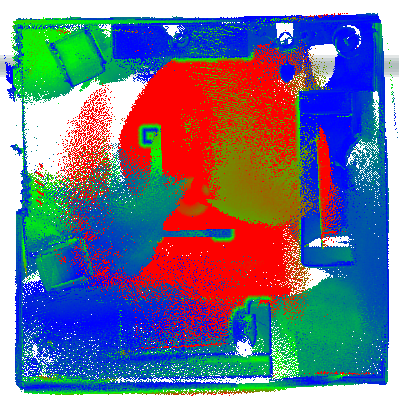}\caption{ORB-SLAM2 Odometry}\label{fig:tabc}\end{subfigure}&
\begin{subfigure}{0.25\textwidth}\centering\includegraphics[width=1\linewidth]{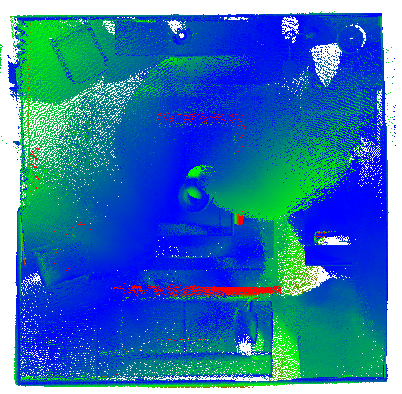}\caption{Full ORB-SLAM2}\label{fig:taba}\end{subfigure}&
\begin{subfigure}{0.25\textwidth}\centering\includegraphics[width=1\linewidth]{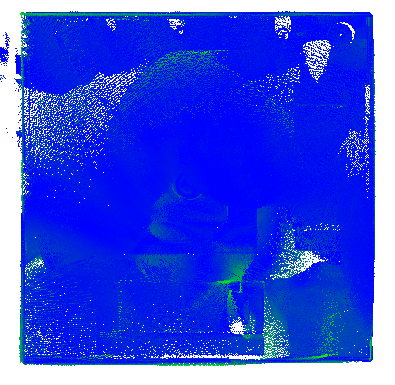}\caption{Ours}\label{fig:taba2}\end{subfigure}\\
\end{tabular}
\caption{Distance error map of reconstructed models from different methods against ground-truth on Living room 1 data sequence.}
\label{fig:error_maps}
\end{figure*}
}

\section{RESULTS}
\label{sec:res}
We have evaluated our LoopSmart system on different datasets and compare it with other online and offline methods. When possible, results reported by authors of original papers are used.

\subsection{Synthetic scenes}

We use augmented ICL-NUIM dataset~\cite{Choi_2015_CVPR} to quantitatively analysis performance of our system. This dataset is a synthetic dataset with ground-truth surface model and camera trajectory. It has two indoor scenes: a living room and an office, and four RGB-D sequences. 

We compare performance of our LoopSmart system with other systems, including ElasticFusion~\cite{whelan2015elasticfusion}, BundleFusion~\cite{Dai_bundle_fusion} and CZK~\cite{Choi_2015_CVPR}. Since we use tracking module from ORB-SLAM2~\cite{mur_artal_orb_slam2} system, we also compare with ORB-SLAM2 odometry part and the full ORB-SLAM2 system. Evaluation metrics are camera trajectory translation RMSE described by Handa \etal and the mean and median distance of the reconstructed surfaces to the ground-truth surfaces in the same way as Whelan \etal. We report them separately in Table \ref{tb:traj_error} and Table \ref{tb:surface_error}. Since different system uses different ways to fuse 3D model, for a fare comparison, we fuse 3D models using ElasticFusion using same parameters with camera trajectory estimate from each system. We use truncating depth distance of $4$ meter and $10$ as surfel confidence threshold for fusion. Specially, for ElasticFusion, we use truncating distance of $5$ meter to for better camera tracking.  

{\centering
\begin{figure}[ht]
\begin{tabular}{ccc}
\begin{subfigure}{0.215\textwidth}\centering\includegraphics[width=1\linewidth]{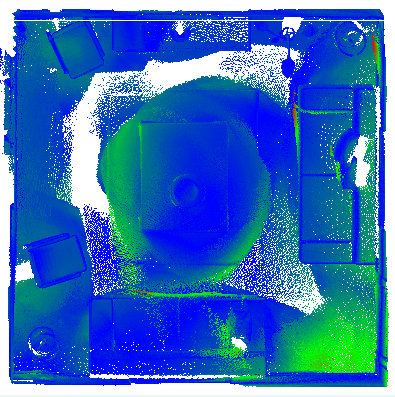}\caption{Full ORB-SLAM2}\label{fig:taba}\end{subfigure}&
\begin{subfigure}{0.215\textwidth}\centering\includegraphics[width=1\linewidth]{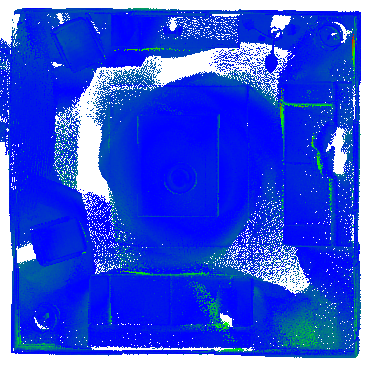}\caption{Ours}\label{fig:tabb}\end{subfigure}&
 \hspace{-0.75cm} \begin{tabular}[c]{@{}l@{}@{}}$\: 0.01\:m$ \\ \begin{subfigure}{0.05\textwidth}\centering\includegraphics[width=0.25\linewidth]{figures/color_scale.png}\label{fig:taba}\end{subfigure}\\ $\quad 0 \: m$\\ $\quad$ \end{tabular}\\
\newline
\begin{subfigure}{0.215\textwidth}\centering\includegraphics[width=1\linewidth]{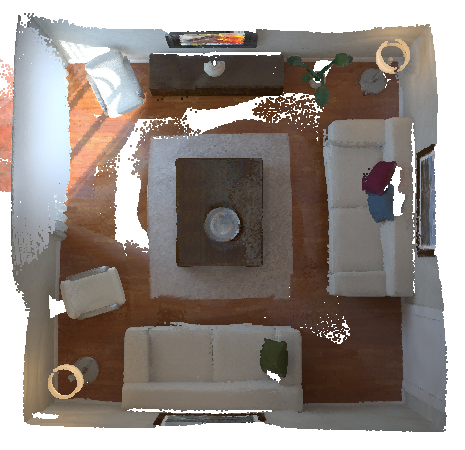}\caption{Full ORB-SLAM2}\label{fig:tabc}\end{subfigure}&
\begin{subfigure}{0.215\textwidth}\centering\includegraphics[width=1\linewidth]{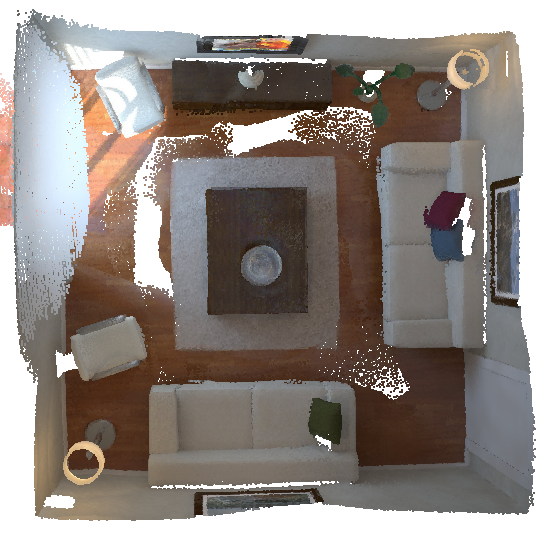}\caption{Ours}\label{fig:taba2}\end{subfigure}&\\
\end{tabular}
\caption{Error map and reconstructed model comparison on Living room 2 data sequence.}
\label{fig:lv2_error}
\end{figure}
}

\begin{table}[h]
\centering
\caption{RMSE error (in meters) of estimated camera trajectories}
\label{tb:traj_error}
\begin{tabular}{|l||c|c|c|c|}
\hline
                                                             & Livingroom 1   & Livingroom 2   & Office 1       & Office 2       \\ \hline \hline
ElasticFusion                                                & 0.50          & 0.36          & 0.14          & 0.10          \\ \hline
BundleFusion                                                 & 1.26          & 0.64          & 2.01          & 0.12          \\ \hline
CZK                                                        & 0.10          & 0.13          & 0.06          & 0.07          \\ \hline
\begin{tabular}[c]{@{}l@{}}ORB-SLAM2 \\ odometry\end{tabular} & 0.14          & 0.05          & 0.05          & 0.03          \\ \hline
\begin{tabular}[c]{@{}l@{}}Full\\ORB-SLAM2\end{tabular}     & 0.10          & 0.03          & 0.04          & 0.03          \\ \hline
Ours                                                         & \textbf{0.04} & \textbf{0.02} & \textbf{0.03} & \textbf{0.02} \\ \hline
\end{tabular}
\end{table}

From Table \ref{tb:surface_error} and Table \ref{tb:traj_error}, our system can give best results on all data sequence in terms of both trajectory and surface estimation accuracy. To give a more informative comparison, we report error map of reconstructed model in Fig. \ref{fig:error_maps} on Living room~1 data sequence. We can see our result have lowest error across the whole model. Other than Living room~1 sequence, ORB-SLAM2 performs also very well. To better understands the difference inside these cases, we report error map and reconstructed model in Fig. \ref{fig:lv2_error} on Living room~2 data sequence. In Fig. \ref{fig:lv2_error}, even though two reconstructed models looks very similar, but the error maps shows our system produce more accurate result.

\subsection{Real-world scenes}

\begin{figure}[ht]
\begin{tabular}{cc}
\begin{subfigure}{0.23\textwidth}\centering\includegraphics[width=1\linewidth]{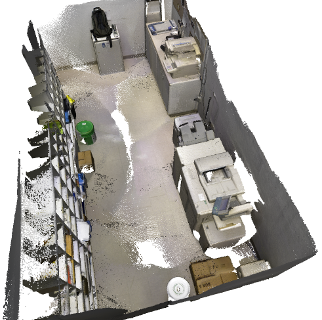}\caption{Copyroom}\label{fig:taba}\end{subfigure}&
\begin{subfigure}{0.23\textwidth}\centering\includegraphics[width=1\linewidth]{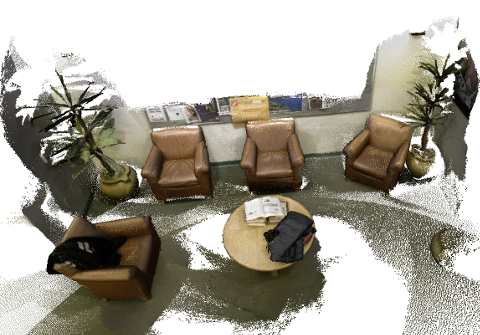}\caption{Lounge}\label{fig:tabb}\end{subfigure}\\
\newline
\begin{subfigure}{0.23\textwidth}\centering\includegraphics[width=2\linewidth]{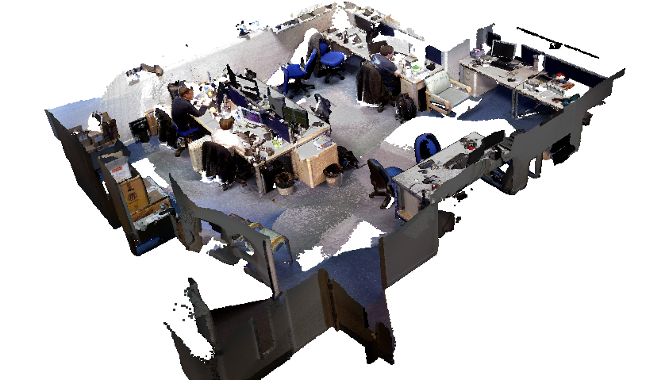}\caption{DysonLab}\label{fig:tabc}\end{subfigure}&
 \\
\end{tabular}
\caption{Reconstructed models of real-world scenes.}
\label{fig:real_world}
\end{figure}

We run experiments on public real-world datasets for a qualitative analysis and a robustness test. We run our system on Copyroom and Lounge dataset from Zhou \etal and DysonLab dataset from Whelan \etal. We report our qualitative results in Fig. \ref{fig:real_world}. Since there are no ground-truth models available, we only report screen-shot of reconstructed model from our system. Due to space limitation, we do not include results of other systems. Interested reader can refer to authors release. Visually, our system produces results at least matching the state-of-the-art methods.

\begin{figure}[H]
\begin{center}
   \includegraphics[width=0.5\linewidth]{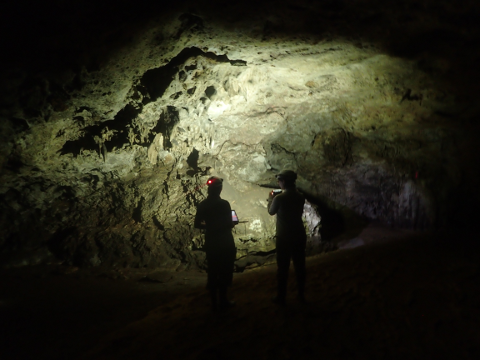}
\end{center}
   \caption{Data being collected in caves.}
\label{fig:cave_data_collecting}
\end{figure}

We also run experiments on a dataset collected by a team of archaeologists. This dataset is collected in Maya caves using a RGB-D sensor Kinect V1 at Belize. LED lights are used to light up the environment as in Fig. \ref{fig:cave_data_collecting} and caves are scanned with loopy motion. We compared results from ORB-SLAM2, ElasticFusion and our system. Note that we set ICP weight to 80 for ElasticFusion to get better tracking. The ORB-SLAM2 suffers from closing loops due to lighting changes and appearance similarity across frames. We show typical results in Fig. \ref{fig:cave_res}. Our system can produce more consistent 3D models, thus get better performance than other systems. When compared with ORB-SLAM2, it shows the importance of closing dense loops. Compared with ElasticFusion, it shows the advantages of having global registration and bundle adjustment as the way to optimize loops.

{\centering
\begin{figure*}[ht]
\begin{tabular}{ccc}
\begin{subfigure}{0.28\textwidth}\centering\includegraphics[width=1\linewidth]{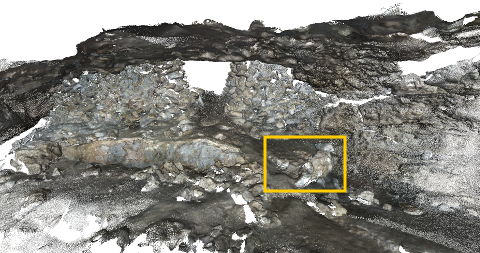}\caption{Ours}\label{fig:taba}\end{subfigure}&
\begin{subfigure}{0.28\textwidth}\centering\includegraphics[width=1\linewidth]{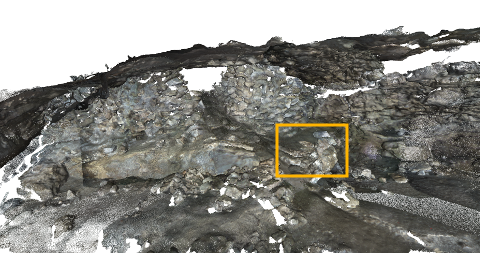}\caption{Full ORB-SLAM2}\label{fig:taba}\end{subfigure}&
\begin{subfigure}{0.28\textwidth}\centering\includegraphics[width=1\linewidth]{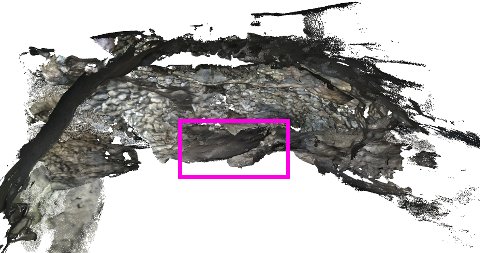}\caption{ElasticFusion}\label{fig:tabb}\end{subfigure}\\
\newline
\begin{subfigure}{0.28\textwidth}\centering\includegraphics[width=1\linewidth]{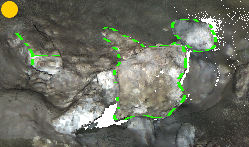}\caption{Ours}\label{fig:tabc}\end{subfigure}&
\begin{subfigure}{0.28\textwidth}\centering\includegraphics[width=1\linewidth]{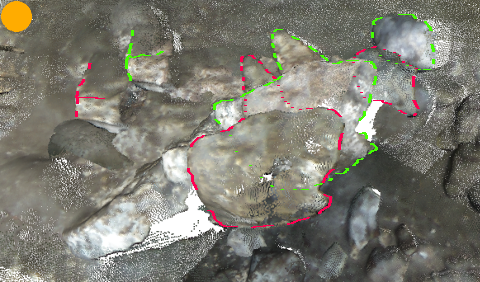}\caption{Full ORB-SLAM2}\label{fig:taba}\end{subfigure}&
\begin{subfigure}{0.28\textwidth}\centering\includegraphics[width=1\linewidth]{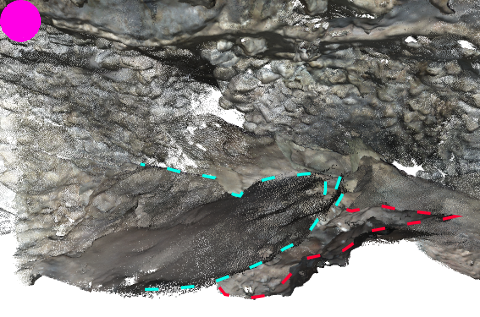}\caption{ElasticFusion}\label{fig:taba2}\end{subfigure}\\
\end{tabular}
\caption{Visualization of reconstructed dense models from different methods. We add reference lines to help readers identity mismatches. For example, in Fig. 6(e), the surface circled in red should be matched to the surface circled in green.}
\label{fig:cave_res}
\end{figure*}
}

\section{CONCLUSIONS}
\label{sec:conclude}
We demonstrate that the LoopSmart framework, proposed of combining dense loops with sparse feature optimization, delivers state-of-art performance. We also want to emphasize that our dense visualization part is optional, so it is possible to further improve performance to make it available on an embedded system, such as an NVIDIA TX 2.

\section{FUTURE WORK}

We target this paper at the framework level innovation to introduce dense loop into sparse feature based tracking and optimization SLAM system, so we put together a working system which enable us to test this framework. It is possible to improve submodules to further improve final results. For example, the tracking module can combine both sparse feature matching and dense ICP based approach to improve robustness at feature-less area. The dense loop verification module can be improved by add hard constraints of angular and positional movement during optimization. We put them in our future work to further improve our system.






\section*{ACKNOWLEDGMENT}

We thank Mighty Chen and Kevin Ong for their help in converting format of RGB-D data.

\bibliographystyle{IEEEtran}
\bibliography{IEEEabrv,egbib}

\begin{thebibliography}{10}
\providecommand{\url}[1]{#1}
\csname url@rmstyle\endcsname
\providecommand{\newblock}{\relax}
\providecommand{\bibinfo}[2]{#2}
\providecommand\BIBentrySTDinterwordspacing{\spaceskip=0pt\relax}
\providecommand\BIBentryALTinterwordstretchfactor{4}
\providecommand\BIBentryALTinterwordspacing{\spaceskip=\fontdimen2\font plus
\BIBentryALTinterwordstretchfactor\fontdimen3\font minus
  \fontdimen4\font\relax}
\providecommand\BIBforeignlanguage[2]{{%
\expandafter\ifx\csname l@#1\endcsname\relax
\typeout{** WARNING: IEEEtran.bst: No hyphenation pattern has been}%
\typeout{** loaded for the language `#1'. Using the pattern for}%
\typeout{** the default language instead.}%
\else
\language=\csname l@#1\endcsname
\fi
#2}}

\bibitem{xiao2013sun3d}
J.~Xiao, A.~Owens, and A.~Torralba, ``{SUN3D}: A database of big spaces
  reconstructed using {SfM} and object labels,'' in \emph{Proc. of the 2013
  IEEE International Conference on Computer Vision}, Dec 2013, pp. 1625--1632.

\bibitem{dai2017scannet}
A.~Dai, A.~X. Chang, M.~Savva, M.~Halber, T.~Funkhouser, and M.~Nie{\ss}ner,
  ``{ScanNet}: {R}ichly-annotated {3D} {R}econstructions of {I}ndoor
  {S}cenes,'' in \emph{Proc. Computer Vision and Pattern Recognition (CVPR),
  IEEE}, 2017.

\bibitem{gzhangicuas2017}
G.~Zhang, B.~Shang, Y.~Chen, and H.~Moyes, ``{SmartCaveDrone}: {3D} cave
  mapping using {UAVs} as robotic co-archaeologists,'' in \emph{Proc. of the
  2017 International Conference on Unmanned Aircraft Systems (ICUAS)}, June
  2017, pp. 1052--1057.

\bibitem{Cadena16tro-SLAMfuture}
C.~Cadena, L.~Carlone, H.~Carrillo, Y.~Latif, D.~Scaramuzza, J.~Neira, I.~Reid,
  and J.~Leonard, ``Past, present, and future of simultaneous localization and
  mapping: Towards the robust-perception age,'' \emph{{IEEE Transactions on
  Robotics}}, vol.~32, no.~6, p. 1309–1332, 2016.

\bibitem{7807268}
D.~Lefloch, M.~Kluge, H.~Sarbolandi, T.~Weyrich, and A.~Kolb, ``Comprehensive
  use of curvature for robust and accurate online surface reconstruction,''
  \emph{IEEE Transactions on Pattern Analysis and Machine Intelligence},
  vol.~PP, no.~99, pp. 1--1, 2017.

\bibitem{Whelan12rssw}
T.~Whelan, M.~Kaess, M.~Fallon, H.~Johannsson, J.~Leonard, and J.~McDonald,
  ``Kintinuous: Spatially extended {K}inect{F}usion,'' in \emph{Proc. of RSS
  Workshop on RGB-D: Advanced Reasoning with Depth Cameras}, Sydney, Australia,
  Jul 2012.

\bibitem{endres20143}
F.~Endres, J.~Hess, J.~Sturm, D.~Cremers, and W.~Burgard, ``3-{D} mapping with
  an {RGB-D} camera,'' \emph{IEEE Transactions on Robotics}, vol.~30, no.~1,
  pp. 177--187, 2014.

\bibitem{mur_artal_orb_slam2}
R.~Mur-Artal and J.~D. Tardos, ``{ORB-SLAM2}: {A}n {O}pen-{S}ource {SLAM}
  {S}ystem for {M}onocular, {S}tereo, and {RGB-D} {C}ameras,'' \emph{IEEE
  Transactions on Robotics}, vol.~PP, no.~99, pp. 1--8, 2017.

\bibitem{GalvezTRO12}
D.~G\'alvez-L\'opez and J.~D. Tard\'os, ``Bags of binary words for fast place
  recognition in image sequences,'' \emph{IEEE Transactions on Robotics},
  vol.~28, no.~5, pp. 1188--1197, October 2012.

\bibitem{newcombe2011kinectfusion}
R.~A. Newcombe, S.~Izadi, O.~Hilliges, D.~Molyneaux, D.~Kim, A.~J. Davison,
  P.~Kohi, J.~Shotton, S.~Hodges, and A.~Fitzgibbon, ``{KinectFusion}:
  Real-time dense surface mapping and tracking,'' in \emph{Proc. of the 2011
  10th IEEE International Symposium on Mixed and Augmented Reality}, Oct 2011,
  pp. 127--136.

\bibitem{whelan2015elasticfusion}
T.~Whelan, S.~Leutenegger, R.~S. Moreno, B.~Glocker, and A.~Davison,
  ``{ElasticFusion}: Dense {SLAM} {W}ithout {A} {P}ose {G}raph,'' in
  \emph{Proc. of Robotics: Science and Systems}, Rome, Italy, July 2015.

\bibitem{990501}
A.~J. Davison and N.~Kita, ``{3D} simultaneous localisation and map-building
  using active vision for a robot moving on undulating terrain,'' in
  \emph{Proceedings of the 2001 IEEE Computer Society Conference on Computer
  Vision and Pattern Recognition. CVPR 2001}, vol.~1, 2001, pp. I--384--I--391
  vol.1.

\bibitem{1241885}
M.~Montemerlo and S.~Thrun, ``Simultaneous localization and mapping with
  unknown data association using {FastSLAM},'' in \emph{Proc. of the 2003 IEEE
  International Conference on Robotics and Automation (Cat. No.03CH37422)},
  vol.~2, Sept 2003, pp. 1985--1991 vol.2.

\bibitem{4209727}
R.~O. Castle, D.~J. Gawley, G.~Klein, and D.~W. Murray, ``Towards simultaneous
  recognition, localization and mapping for hand-held and wearable cameras,''
  in \emph{Proceedings 2007 IEEE International Conference on Robotics and
  Automation}, April 2007, pp. 4102--4107.

\bibitem{Dai_bundle_fusion}
A.~Dai, M.~Niessner, M.~Zollhofer, S.~Izadi, and C.~Theobalt, ``{BundleFusion}:
  Real-time globally consistent {3D} reconstruction using on-the-fly surface
  reintegration,'' \emph{ACM Trans. Graph.}, vol.~36, no.~3, pp. 24:1--24:18,
  May 2017.

\bibitem{keller2013real}
M.~Keller, D.~Lefloch, M.~Lambers, S.~Izadi, T.~Weyrich, and A.~Kolb,
  ``Real-time 3{D} reconstruction in dynamic scenes using point-based fusion,''
  in \emph{Proc. of the 2013 International Conference on 3D Vision-3DV
  2013}.\hskip 1em plus 0.5em minus 0.4em\relax IEEE, 2013, pp. 1--8.

\bibitem{Choi_2015_CVPR}
S.~Choi, Q.~Y. Zhou, and V.~Koltun, ``Robust reconstruction of indoor scenes,''
  in \emph{Proc. of the 2015 IEEE Conference on Computer Vision and Pattern
  Recognition (CVPR)}, June 2015, pp. 5556--5565.

\bibitem{zhou2016fast}
Q.-Y. Zhou, J.~Park, and V.~Koltun, ``Fast {G}lobal {R}egistration,'' in
  \emph{Proc. of the European Conference on Computer Vision}.\hskip 1em plus
  0.5em minus 0.4em\relax Springer, 2016, pp. 766--782.

\bibitem{5152473}
R.~B. Rusu, N.~Blodow, and M.~Beetz, ``Fast point feature histograms ({FPFH})
  for {3D} registration,'' in \emph{Proc. of the 2009 IEEE International
  Conference on Robotics and Automation}, May 2009, pp. 3212--3217.

\bibitem{Fischler:1981:RSC:358669.358692}
M.~A. Fischler and R.~C. Bolles, ``Random sample consensus: A paradigm for
  model fitting with applications to image analysis and automated
  cartography,'' \emph{Commun. ACM}, vol.~24, no.~6, pp. 381--395, June 1981.

\end{thebibliography}

\end{document}